\pdfoutput=1 
\documentclass[journal]{IEEEtran}

\IEEEoverridecommandlockouts                              % This command is only needed if 
\usepackage{cite}
\usepackage{graphics} % for pdf, bitmapped graphics files
\usepackage{epsfig} % for postscript graphics files
\usepackage{epstopdf}
\graphicspath {{./image/}}
\usepackage{mathptmx} % assumes new font selection scheme installed
\usepackage{times} % assumes new font selection scheme installed
\usepackage{amsmath} % assumes amsmath package installed
\usepackage{amssymb}  % assumes amsmath package installed
\usepackage{gensymb}
\usepackage{multirow}
\usepackage{picinpar}
\usepackage{multicol}
\usepackage{stfloats}
\usepackage{caption}
\usepackage{graphicx}
\usepackage{subfigure}
\usepackage{wasysym}
\usepackage{textcomp}
\usepackage{paralist}
\usepackage{scalerel,stackengine}
\stackMath

\usepackage{xargs}                      % Use more than one optional parameter in a new commands
\usepackage[pdftex,dvipsnames]{xcolor}  % Coloured text etc.
\usepackage[colorinlistoftodos,prependcaption,textsize=tiny]{todonotes}
\newcommandx{\sen}[2][1=]{\todo[inline,linecolor=green,backgroundcolor=green!25,bordercolor=blue,#1]{\textbf{Sen}: #2}}

\begin{document}

\title{SGANVO: Unsupervised Deep Visual Odometry and Depth
Estimation with Stacked Generative Adversarial Networks}%From Videos}

\author{Tuo Feng$^{1}$ and Dongbing Gu$^{1}$% <-this % stops a space
\thanks{Tuo Feng and Dongbing Gu are with the School of Computer Science and Electronic Engineering, University of Essex, Colchester, CO4 3SQ, UK. {\tt\small \{tfeng, dgu\}@essex.ac.uk}}%% <-this % stops a space
\thanks{}% <-this % stops a space
\thanks{}}

%\markboth{IEEE Transactions on Cognitive and Developmental Systems,~Vol.~XXX, No.~XXX, XXX~2018}%
%{Shell \MakeLowercase{\textit{et al.}}: Bare Demo of IEEEtran.cls for Journals}
\maketitle

\begin{abstract}
Recently end-to-end unsupervised deep learning methods have achieved an effect beyond geometric methods for visual depth and ego-motion estimation tasks. These data-based learning methods perform more robustly and accurately in some of the challenging scenes. The encoder-decoder network has been widely used in the depth estimation and the RCNN has brought significant improvements in the ego-motion estimation. Furthermore, the latest use of Generative Adversarial Nets (GANs) in depth and ego-motion estimation has demonstrated that the estimation could be further improved by generating pictures in the game learning process. This paper proposes a novel unsupervised network system for visual depth and ego-motion estimation: Stacked Generative Adversarial Network (SGANVO). It consists of a stack of GAN layers, of which the lowest layer estimates the depth and ego-motion while the higher layers estimate the spatial features. It can also capture the temporal dynamic due to the use of a recurrent representation across the layers. See Fig. 1 for details. We select the most commonly used KITTI\cite{c1} data set for evaluation. The evaluation results show that our proposed method can produce better or comparable results in depth and ego-motion estimation.
\end{abstract}

%%%%%%%%%%%%%%%%%%%%%%%%%%%%%%%%%%%%%%%%%%%%%%%%%%%%%%%%%%%%%%%%%%%%%
%\begin{IEEEkeywords}
%SLAM, unsupervised deep learning, depth estimation, RCNN, machine learning.
%\end{IEEEkeywords}

\IEEEpeerreviewmaketitle

%%%%%%%%%%%%%%%%%%%%%%%%%%%%%%%%%%%%%%%%%%%%%%%%%%%%%%%%%%%%%%%%%%%%%
\section{INTRODUCTION}
Object's depth estimation and camera's ego-motion estimation are two essential tasks in autonomous robotic applications. Impressive progress has been achieved by using various geometric based methods. Recently, unsupervised deep learning based methods have demonstrated a certain level of robustness and accuracy in some challenging scenes without the need of labeled training data set.

	\begin{figure}
		\centering
		\begin{minipage}[a]{1.0\columnwidth}
         \includegraphics[width=1.0\columnwidth]{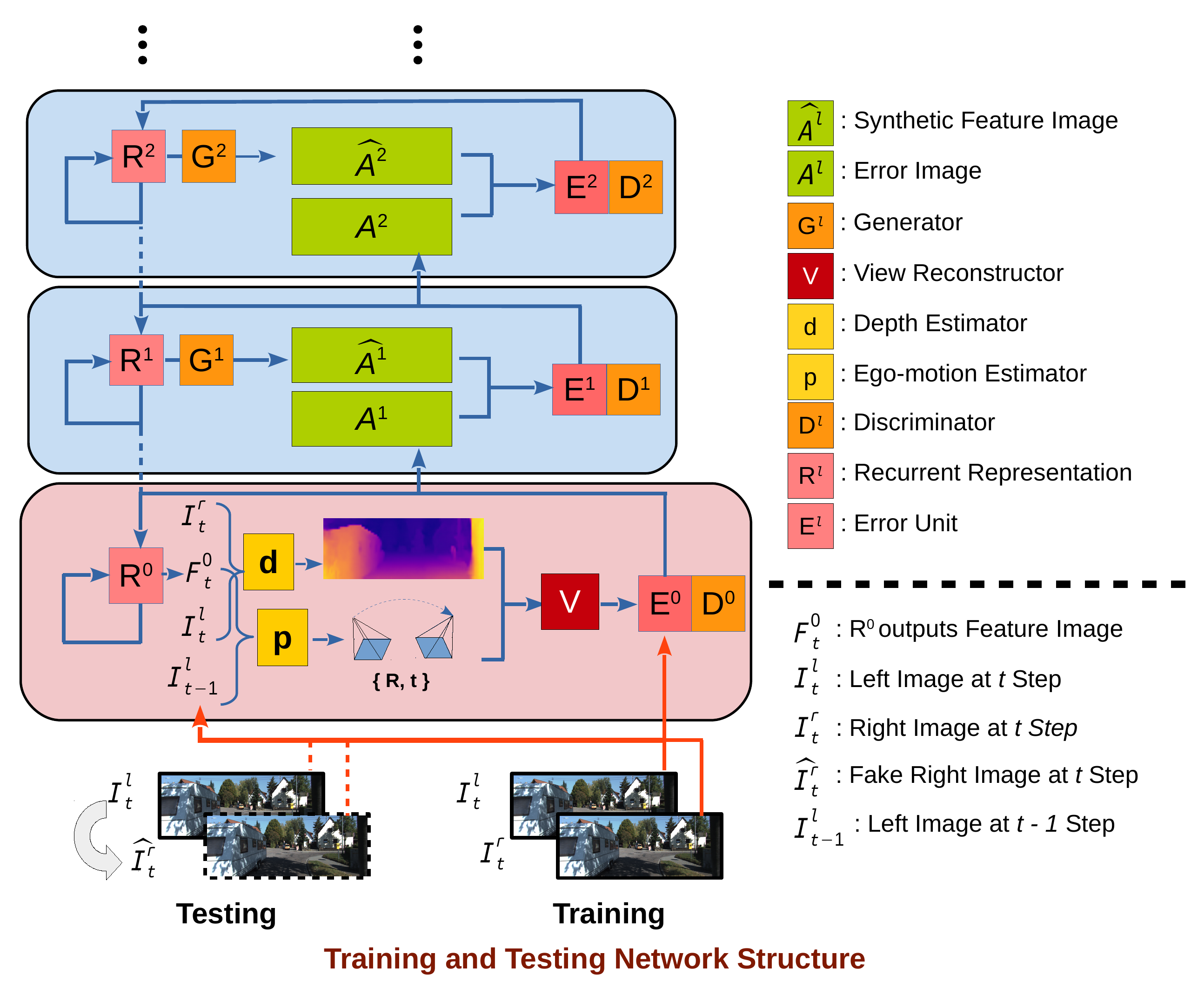}
		\caption{Our proposed SGANVO architecture for the depth and ego-motion estimation. It is a series of stacked GANs. The bottom layer estimates the depth and ego-motion. The other layers estimate the spatial features.}
		\label{net_structure}
		 \end{minipage}
	\end{figure}

Inspired by the image wrapping
technique - spatial transformer\cite{c2}, Garg et al.\cite{c3} proposed an unsupervised Convolutional Neural Network (CNN) to estimate the depth by using the left-right photo-metric constraint of stereo image pairs. Godard\cite{c4} further extended this method by employing a loss function with the left and right images wrapping  across each other. Because both left and right photo-metric losses are penalized, this method improved the accuracy of depth estimation. Zhou et al.\cite{c5} proposed two separate networks to infer the depth and ego-motion estimation over three temporal consecutive monocular image frames. The middle frame performs as the target frame and the previous and the following frames as the source frames. Yin\cite{c6} proposed the GeoNet which not only estimates the depth and ego-motion but also the optical flow for dynamic objects. Similarly, Godard et al.\cite{c7} proceeded to merge the pose network with the depth network through sharing the network weights. Furthermore, benefit from the recent advance of deep learning methods for single-image super-resolution, Pillai et al.\cite{c8} proposed the sub-pixel convolutional layer extension for obtaining depth super-resolution. Their superior pose network is bootstrapped with much more accurate depth estimation. 

	\begin{figure}
		\centering
		\begin{minipage}[a]{1.0\columnwidth}
         \includegraphics[width=1.0\columnwidth]{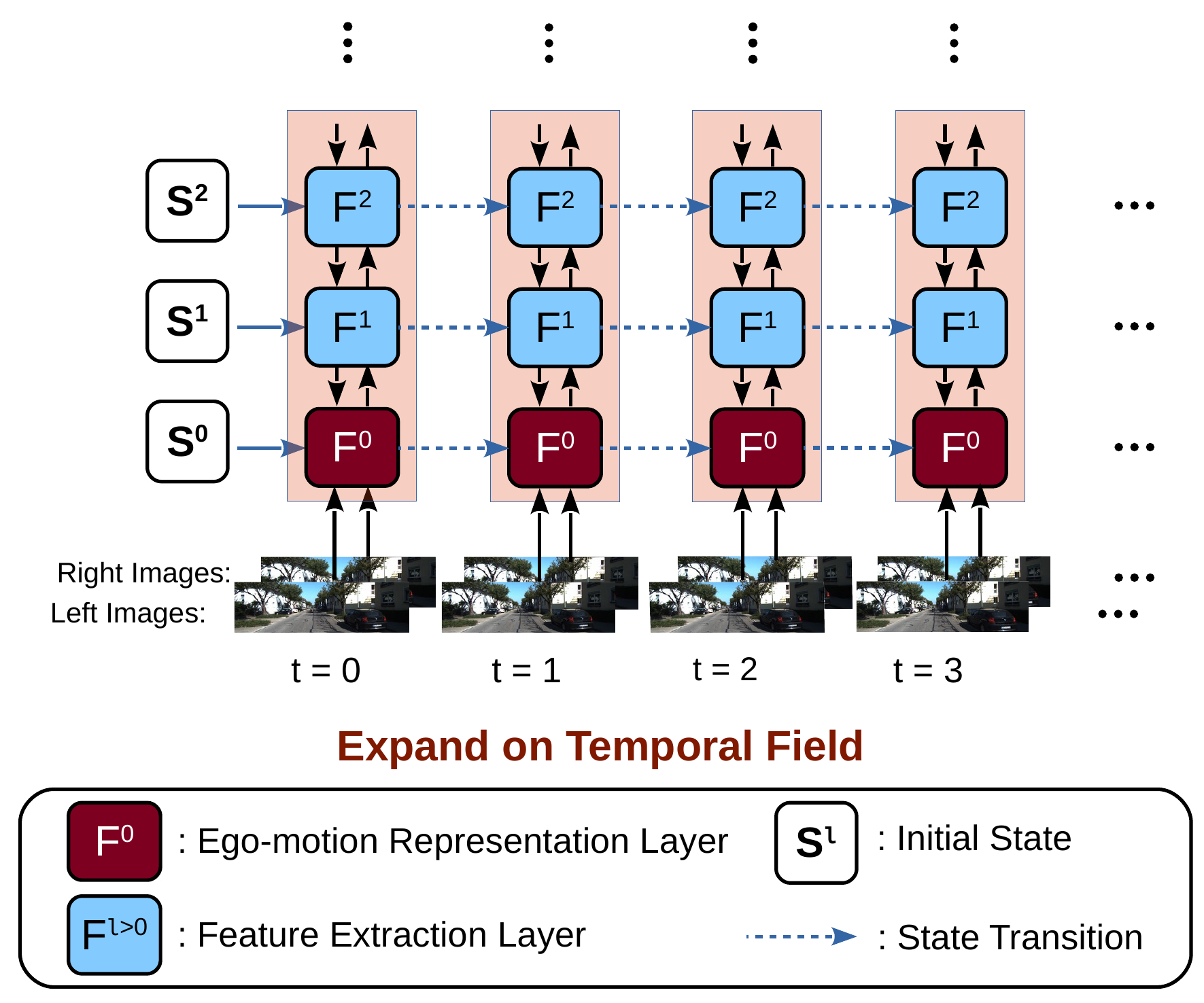}
		\caption{The network is unfolded in time. The temporal dynamic is captured in the recurrent representation.}
		\label{net_ext_time}
		 \end{minipage}
	\end{figure}

Most of recent work are all based on the encoder-decoder neural networks, but with different loss functions based on view reconstruction approach. Li et al.\cite{c9} designed their loss functions not only combining the loss functions defined in \cite{c4} and \cite{c5} but also adding the 3D points transformation loss. However, their results perform not very well in the scenes where dynamic and occluded objects occupy a large part of the field of view due to artificially designed rigid transformation used. To deal with dynamic and occluded objects in the scenes, Ranjan \cite{c10} and Wang \cite{c11} introduced additional networks to estimate the optical flow jointly with the depth and pose networks from videos. Ranjan used two networks to learn the optical flow and dynamic object masks, and jointly compute the flow and mask loss functions to refine the depth and pose networks. Wang added the optical flow estimation network to deal with the dynamic objects and refine the depth network. Due to the use of a PWC network\cite{c12} to handle the stereo depth estimation, Wang's Undepthflow improved the unsupervised depth estimation by an order of magnitude and its pose estimation also reached the optimal rank. Almalioglu\cite{c13} introduced a visual odometry (GANVO) to estimate the depth map using Generative Adversarial Network\cite{c14}. Its Generator network generates the depth map and the pose regressor infers the ego-motion. Together they build up the view reconstruction. Then its Discriminator network games the original target image and the reconstructed image to refine the depth and pose networks. This work is similar with \cite{c15} \cite{c16} \cite{c17} \cite{c18}, which use the Generator to generate the depth map instead of using estimation networks. From these work, it can be seen that the generative nature in GAN networks is beneficial to the scenes with dynamic objects. However, these visual odometry related GANs focus on the depth estimation, but not on the ego-motion estimation. 

In this paper, we propose a novel unsupervised deep visual odometry system with Stacked Generative Adversarial Networks (SGANVO) (see Fig. 1). Our main contributions are as follows:
\begin{itemize}

\item To the best of our knowledge, this is the first time to use the stacked generative and adversarial learning approach for joint ego-motion and depth map estimation.
\item Our learning system is based on a novel unsupervised GAN scheme without the need for ground truth.
\item Our system possesses a recurrent representation which can capture the temporal dynamic features.
%\item Based on the improved Wasserstein GAN (WGAN)\cite{gulrajani2017improved}, our SGANVO not only alleviates the problem of unstable training, but also provides a reliable indicator of the training progress.

\end{itemize}

%Contrary to GANVO, our SGANVO is fed with stereo imagery for training without the need of labeled data-sets. The result of the stereo imagery training can recover the real scale from monocular image input. Potentially, it is possible to train our SGANVO with an extremely massive number of unlabeled data-sets, and expand the temporal window to continuously improve its performance.

The outline of this paper is organized as follows: Section II gives an overview of our proposed SGANVO system. Section
III describes various loss functions used for the Generator and Discriminator. Section IV presents our experimental results of depth and ego-motion estimation. Finally, conclusion and future work are drawn in Section V.

%%%%%%%%%%%%%%%%%%%%%%%%%%%%%%%%%%%%%%%%%%%%%%%%%%%%%%%%%%%%%%%%%%%%%
\section{SYSTEM OVERVIEW}
We address the depth and the ego-motion estimation as a whole visual odometry system in Fig.1. The system consists of an ego-motion representation layer for ego-motion estimation, and multiple feature extraction layers for feature estimation. The error image $A^l$ from error unit $E^{l-1}$ in current layer is propagated up to higher layer as the input. The hidden state in recurrent representation layer $R^l$ is propagated down to lower layer for capturing the temporal dynamic. There are mainly two units, Generator ($G^l$) and Discriminator ($D^l$), in each layer.

Fig.2 shows the network unfolding in time step. After initial states $S^l$ are set to the network. A current frame is fed to the network input. This is run for a sequence of $N$ consecutive frames. After $N$ steps, The initial states are set to the network again.

\subsection{Generator}
In the bottom layer, the Generator consists of depth estimator $d$, ego-motion estimator $p$, and view reconstructor $V$. It takes the hidden state from recurrent representation $R^0$ as its input and generates an estimated image from view reconstructor $V$ for current input image. In the higher layer ($l>0$), the Generator is a convolutional layer which  takes the hidden state from recurrent representation $R^l$ as its input and generates an estimated error image $\hat{A}^l$ for the input $A^l$ from lower layer. The recurrent representation $R^l$ is a Long Short Term Memory network (LSTM) and connected top down between layers. The hidden state $R^l_t$ is updated according to $R^l_{t-1}, R^{l+1}_t, E^l_{t-1}$. It is used to store the temporal dynamic from consecutive video which is defined as:
\begin{equation}
	R^l_t = ConvLSTM( E^l_{t-1}, R^l_{t-1}, UpSample(R^{l+1}_t) )
	\end{equation}
The hidden state $R^l_t$ is updated through two passes: a top down pass from the \textit{UpSample} of $R^{l + 1}$, and then a level pass calculated by previous hidden state $R^{l}_{t-1}$  and previous error $E^{l}_{t-1}$.

The depth estimator is an encoder-decoder network to generate the multi-scale dense depth map, like the one used in \cite{c4}. Inspired by \cite{c8}, we replace the \textit{UpSample} branches in the depth decoder with the sub-pixel convolutional branches used in \cite{c19} to generate the up-scaled features. Sub-pixel convolutional branches consist of a sequence of three consecutive 2D convolutional layers with 64, 32, 4 output channels with 1 pixel stride.
The depth estimator directly generates the dense depth map by using a pair of stereo images separately concatenating the features from unit $R^0_t$ along with the image channel to train the network.  

The ego-motion estimator is a VGG-based CNN architecture. It takes current image and the hidden state representing the information in previous frame as the input, and generates the 6-DoF ego-motion estimation between current frame and previous frame. We decouple the translation and the rotation with two separate groups of fully-connected layers after the last convolutional layer for better performance. We feed the input data concatenating two continuous images and the features from unit $R^0_t$ along with image channel to train the ego-motion estimator. 

%Because of inputting one frame each time, the pose estimator will generate a 6-DoF ego-motion for this frame like \cite{wang2018recurrent}. However, the first frame of each temporal window will introduce the failure initialization of pose estimation. To resolve that, we reuse the first frame as the initial last frame input and start to compute pose translation from the second frame.

%at the bottom of the stack takes one stereo image at each time step within a consecutive temporal window. Then it computes the error between reconstruction and original frame and feed to the high-level Generators and the bottom feed yielding layer. The feed yielding layer produces the dynamic feature to respectively ego-motion and depth estimators to generate the scaled 6-DoF ego-motion and depth outputs. 

%From the second level, Generator only uses the difference of the original feed image and view reconstructed image from the lower level as this level's feed. Analogously, the higher Generators generate the higher dimensional differences as outputs.

\begin{figure}
		\centering
		\includegraphics[width=1.0\columnwidth]{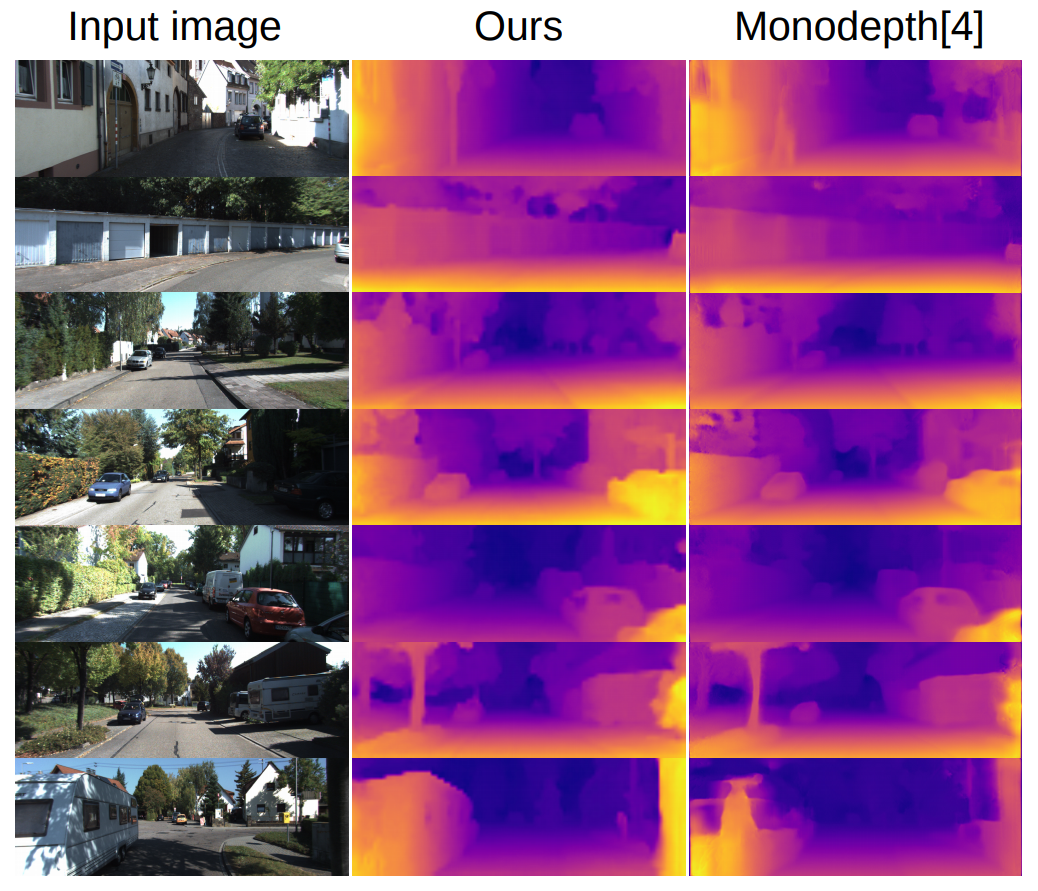}
		\caption{ Illustrated above are qualitative comparisons of our SGANVO (image size: $416\times128$) with Monodepth \cite{c4} (image size: $512\times256$). The depth map shows that our approach produces qualitatively better depth estimates with crisp boundaries.}
		\label{fig:TrainingScheme}
	\end{figure}

The view reconstructor $V$ takes the depth estimate, ego-motion estimate, and previous frame $I_{t-1}$ as the input and generates a predicted image $\hat{I}_t$ for the current frame $I_t$. By using the spatial transformer network \cite{c2} and the homogeneous coordinate of the pixel $I_{t-1}(i,j)$ in the $(t-1)$th frame, we can derive its corresponding pixel $\hat{I}_t(i,j)$ in the $t$th frame through
	\begin{equation}
	\hat{I}_t(i,j) = K\hat{T}_{t-1,t}d_{t-1}^{-1}(i,j)K^{-1}I_{t-1}(i,j)
	\end{equation}
where $K$ is the camera intrinsic matrix, $d_{t}(i,j)$ is the estimated disparity, $\hat{T}_{t-1, t}$ is the camera coordinate transformation matrix from the $(t-1)$th frame to the $t$th frame generated by the ego-motion estimator. 

Based on this, $\hat{I}_{t-1}$ and $\hat{I}_{t}$ can be constructed from ${I}_{t}$ and ${I}_{t-1}$, respectively. 
%Like \cite{godard2017unsupervised} computing the left-right consistency loss, we replace the left to right optical flow with temporal target to source optical flow to synthesize the warping frame. The more details about Generators' losses during training procedure are present in the next section. 
The error units in the bottom layer or higher layers are computed as below: 
    \begin{align}
	E_{t}^{0} &= [ReLU({I}_t-{\hat{I}_{t}});ReLU({\hat{I}_{t}}-{I}_{t})]\\
	E_t^{l} &= [ReLU({A}_{t}^{l}-{\hat{A}_{t}^{l}});ReLU({\hat{A}_{t}^{l}}-{A}_{t}^{l})]
	\end{align}
where $E_t^0$ is the error image between the generated from the view reconstructor and the current frame at $t$ time, $E_t^l$ is the error image at the higher layer ($l>0$), $ReLU$ is an activation function operation. The input error image $A^l_t$ is defined as:
    \begin{align}
	A_t^{l} = Maxpool(ReLU(Conv(E_t^{l -1})))
	\end{align}
where $Conv$ denotes the convolutional operation. The Generator in the higher layer ($l>0$) consists of one convolutional unit with 3 dimension filters to generate error feature image $\hat{A}^l_t$:
\begin{align}
	\hat{A}_t^{l} = ReLU(Conv(R_t^l))
	\end{align}

%Generator synthesizes the new frame through rigid alignment with the previous frame, current frame's generated depth and generated ego-motion. To improve the dynamic objects' estimation accuracy, both depth and ego-motion estimation come from the feed yielding layers temporal features learning and using the synthetic frame to compute Generators' losses from all frames in the temporal window. For target frame $I_{k}$ and source frame $I_{k+1}$ with some scene overlaps, we can obtain their synthesized images ${I}'_{k}$ and ${I}'_{k+1}$ by 

\subsection{Discriminator}
The Discriminator is a convolutional network that can categorize the images fed to it. In the bottom layer, we feed the generated image $\hat{I}_t$ and original image $I_t$ into it. In the higher layer ($l>0$), the inputs to the Discriminator are the generated error image $\hat{A}^l_t$ and the error image $A^{l}_t$ from the lower layer.
We used WGAN in our architecture, which employs the Wasserstein distance instead of Kullback-Leibler (KL) divergence or Jensen-Shannon (JS) divergence for stabilizing the training process using gradient descents. The Discriminator architecture is shown in Table 1. Considering the GPU's memory space, we set all the Discriminators five simple convolutional layers connected by four SELU block units as provided in \cite{c20}. 
\begin{table}%[bp]% [thpb] % [h]
		\centering
		\renewcommand{\arraystretch}{1.45}
		\caption{ Discriminator Network Architecture}
		\label{Discriminator_table}
		%\footnotesize
		% \scriptsize
		\small
		\setlength{\tabcolsep}{9pt}
		\begin{tabular*}{\columnwidth}{c|c|c|c|c}
		\hline
		Block Type & Kernel Size & Strides & Filters & Input\\
		\hline
		Conv1 & $5\times5$ & 2 & 16 & Image\\
		\hline
		SELU1 & none & none & none & Conv1\\
		\hline
		 Conv2 & $5\times5$ & 2 & 32 & SELU1\\
		\hline
		SELU2 & none & none & none & Conv2\\
		\hline
		Conv3 & $5\times5$ & 2 & 64 & SELU2\\
		\hline
			SELU3 & none & none & none & Conv3\\
		\hline
		Conv4 & $5\times5$ & 2 & 128 & SELU3\\
		\hline
		SELU4 & none & none &none  & Conv4\\
		\hline
		Conv5 & $4\times4$ & 1 & 1 & SELU4\\
		\hline
		\end{tabular*}
	\end{table}

	\begin{table*}%[bp]% [thpb] % [h]
		\centering
		\renewcommand{\arraystretch}{3}
		\caption{
		Depth estimation results on KITTI dataset using the split of Eigen et al.\cite{c21}. For
        training data, K = KITTI, CS = Cityscapes \cite{c27}, M = Monocular and S = Stereo. For testing data, our SGANVO uses the monocular dataset like others' work and our view reconstruction component uses this left image sequence to generate the right view image sequence like \cite{c26}.}
		\label{depth_comparison}
		\small
				\centering
		\renewcommand{\arraystretch}{1.45}
		\begin{tabular*}{1.0\textwidth}{c@{\extracolsep{\fill}}ccccccccccc}
	%	\begin{tabular*}{\columnwidth}{ccccccccccc}
		\hline
		Method & Datasize & Dataset & Train & Abs Rel & Sq Rel & RMSE  & RMSE $\log$ &  $\delta$ \textless 1.25 & $\delta$ \textless $1.25^2$  & $\delta$ \textless $1.25^3$ \\ \hline
Garg et al.\cite{c3}              & 620 $\times$ 188 & K       & M     & 0.169   & 1.080  & 5.104 & 0.273    & 0.740            & 0.904            & 0.962             \\

SfMLearner \cite{c5}                     &416 $\times$ 128& K       & M     & 0.208   & 1.768  & 6.856 & 0.283    & 0.678            & 0.885            & 0.957             \\
SfMLearner \cite{c5}                     & 416 $\times$ 128 & CS+K    & M     & 0.198   & 1.836  & 6.565 & 0.275    & 0.718            & 0.901            & 0.960             \\
GeoNet \cite{c6}                         & 416 $\times$ 128 & K       & M     & 0.155   & 1.296  & 5.857 & 0.233    & 0.793            & 0.931            & 0.973             \\
GeoNet \cite{c6}                         & 416 $\times$ 128 & CS+K    & M     & 0.153   & 1.328  & 5.737 & 0.232    & 0.802            & 0.934            & 0.972             \\
Vid2Depth \cite{c22}                      & 416 $\times$ 128 & K       & M     & 0.163   & 1.240  & 6.220 & 0.250    & 0.762            & 0.916            & 0.968             \\
Vid2Depth \cite{c22}                      & 416 $\times$ 128 & CS+K    & M     & 0.159   & 1.231  & 5.912 & 0.243    & 0.784            & 0.923            & 0.970             \\ 	
GANVO \cite{c16}                          & 416 $\times$ 128 & K       & M     & 0.150   & 1.414  & 5.448 & 0.216    & 0.808 & 0.939 &0.975\\ \hline
UnDeepVO \cite{c9}                       & 416 $\times$ 128 & K       & S     & 0.183   & 1.730   & 6.570  & 0.268    & -                & -                & -                 \\
Godard et al.\cite{c4}                   & 640 $\times$ 192 & K       & S     & 0.129   & 1.112  & 5.180 & 0.205    & 0.851            & 0.952            & 0.978             \\
SGANVO                          & 416 $\times$ 128 & K       & S     & \textbf{0.065}   & \textbf{0.673}  & \textbf{4.003} & \textbf{0.136}   & \textbf{0.944}           & \textbf{0.979}            & \textbf{0.991}             \\
\hline

		\end{tabular*}
	\end{table*}

\section{TRAINING PROCEDURE}

The SGANVO training follows the improved WGAN training procedure \cite{c23}. We jointly compute the losses of Generators and Discriminators in all the layers and take the weighted sum as the final G loss and D loss. We feed the stereo data sequence of KITTI dataset into the depth estimator, and the monocular data sequences of KITTI dataset into the ego-motion estimator. The loss functions are defined as below:  

\subsection{Generator Loss} %GANVO \cite{almalioglu2018ganvo} modifies the Generator network for deeper architectures and uses the batch norm in both G and D networks. Then generates non-blurry images and resolves the training convergence problem. This generator's feed solution coincides with ours but there are massive differences. Training's convergence is the main problem to generate the accurate results during applying GAN networks. After a great deal of experimental verification, a random noise or vector cannot generate the corresponding depth or ego-motion successfully even if similar results. Therefore, the Generator's inputs must be the original image or its derivatives. We design the feed yielding layer not only to resolve the convergence problem but also to store the temporal generated error and higher dimensional features. 

Our G losses mainly include three parts. The first one is from the Discriminator and defined as: 
      \begin{equation}      
      L_{g}^{D} = \sum_{l}\lambda_{l}{E[D(\hat{x}^{l})]}
      \end{equation}
where $\lambda_{l}$ is the weight at $l$th layer, and $D(\hat{x}^{l})$ is the output from the Discriminator when its input is from generated image $\hat{x}^{l}$ in its layer. $E$ is the mean operation. 

The second one is the weighted sum of all the error images within a sequence of $N$ consecutive frames.
      \begin{equation}      
      L_{g}^{N} =  \sum_{t=1}^N\lambda_{t}\sum_{l}\frac{\lambda_{l}}{{n}_{l}}\sum_{{n}_{l}} \|E_t^{l}\|_1			
      \end{equation}
where $\lambda_{t}$ is the temporal weight factor, $\lambda_{l}$ is the layer weight factor, $n_{l}$ is the total number of ReLU units of the $l$th layer, and $\|\cdot\|_1$ is the $L_1$ norm.  

The last one is the disparity consistency loss which is used to improve the depth smoothness. Denote $d^{l}_t$ and $d^{r}_t$ the left and right  disparity  maps, respectively. The  disparity  consistency  loss is defined as:
      \begin{equation}      
      L_{g}^{d} =  \sum_{t=1}^{N}\sum_{i,j}\|d^{l}_t(i,j)-{d^{r}_t}(i,j)\|_1 
      \end{equation}

The final G loss is the weighted sum of above three parts. 
 \begin{equation}\label{eq:7}     
      L_{g}^{final} =\alpha L_{g}^{D}+\beta L_{g}^{N}+\gamma L_{g}^{d}
      \end{equation}
      
where $\alpha$, $\beta$ and $\gamma$ are the weight parameters. 

\begin{figure}
		\centering
		\begin{minipage}[a]{1.0\columnwidth}
         \includegraphics[width=1.0\columnwidth]{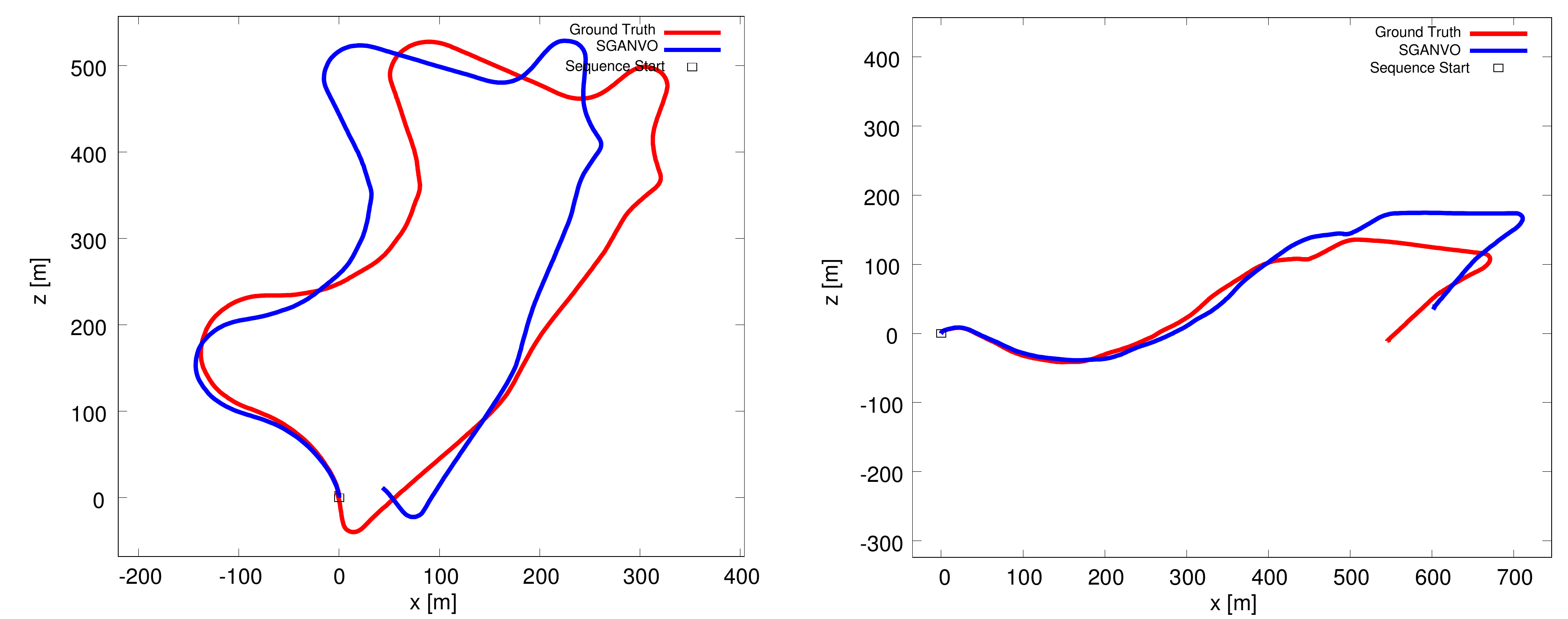}
		\caption{Our proposed SGANVO system to estimate the ego-motion on the KITTI odometry
benchmark, Sequence 09 (left) and Sequence 10 (right). Our results are rendered in blue while the ground truth is rendered in red.}
		\label{depth_images}
		 \end{minipage}
	\end{figure}

\subsection{Discriminator Loss}
The D loss is defined as:
      \begin{equation}      
      L_{d}^{D} = \sum_{l}\lambda_{l}{\arg \max_D D_{d}^{l}}		
      \end{equation}
where $D^{l}_d$ is the Discriminator loss in the $l$th layer when its inputs are real image $x$ and generated image $\hat x$. Afterwards the $D^{l}_d$ is defined as :
      \begin{equation}      
      D_{d}^{l} = {E[{D}(x^l)]-E[{D}(\hat x^{l})]+\lambda_{D}E[(||\triangledown{D}( {x^\prime}^{l})||_2 - 1)^2]}		
      \end{equation}
where $E$ is the mean operation and $\triangledown{}$ is the solving gradient operation. The innovation of WGAN-GP is on the last item of the above loss function where the $x^\prime$ is the random interpolation samples between real value and generated value defined as:
      \begin{equation}      
      x^\prime = x*\epsilon + \hat{x}*(1 - \epsilon)		
      \end{equation}
In our implementation, we choose $\lambda_{D} = 10$ and $\epsilon$ is a random number from uniform distribution between 0 and 1.

\subsection{Adversarial Training Procedure}

To sum up, our SGANVO generates the features of temporal field over a time window and of spatial field over stacked layers. It can estimate the depth and the ego-motion at current time step from previous frame in the bottom layer. 

Initially, the first frame in the first temporal window is fed to the network. This frame is also used as the initial previous frame to resolve the initial matrix irreversible problem. Thus our system can generate one depth estimation and one ego-motion estimation at each time step of the window. We set all the initial states $S^l$ to zeroes. They are passed to the recurrent representation $R^l$. The synthetic error image $\hat{A}^l$ are generated as zero matrices. And the initial estimate of ego-motion is zero as the initial current input and initial previous image are the same. But an initial depth map can be estimated from the depth estimator. 

After the initialization, the first frame is fed to the system, and the system begins to generate the results in each layer. 

After completing the feeding of current temporal window to the network, the system computes all the G the D losses over all the temporal steps ($N$). At the same time, it can generate both the depth and ego-motion estimate. Then the final G and D losses are applied to back propagate the Generators and Discriminators. The min-max training of our SGANVO is described as: 
      \begin{equation}      
      W(x, \hat{x}) = \arg \min_D L_{d}^{D} 	
      \end{equation}
The Discriminator training procedure is to make the $W(x, \hat{x})$ convergent to the minimum. The above procedure is repeated for next temporal window until all the frames in the data set are used. 

\begin{table*}%[bp]% [thpb] % [h]
		\centering
		\renewcommand{\arraystretch}{1.35}
		\caption{Ego-motion estimation results on KITTI dataset with our proposed SGANVO. We also compare with the four learning methods ( Monocular: ESP-VO \cite{c24}, SfMlearner \cite{c5}; Stereo: UndeepVO \cite{c9}, Undepthflow\cite{c11}), and one geometric based methods (ORB-SLAM). Our SGANVO, SfMLearner, and UndeepVO use images with 416$\times$128. ESP-VO and ORB-SLAM use 1242$\times$376 images. Undepthflow uses 832$\times$256 images. The best results among the learning methods are made in bold.}
		\label{RCNN_VO_00-08_table}
		\small
		\begin{tabular*}{1.0\textwidth}{c@{\extracolsep{\fill}}cccccc||cc||cccc}
			\hline
			\multirow{4}{*}{Seq.} & \multicolumn{8}{c||}{Monocular} & \multicolumn{4}{c}{Stereo}\\ \cline{2-13}
			& \multicolumn{2}{c}{SGANVO}    & \multicolumn{2}{c}{ESP-VO \cite{c24}}                   & \multicolumn{2}{c||}{SfMLearner \cite{c5}}                  &
			\multicolumn{2}{c||}{ORB-SLAM \cite{c25}}        & \multicolumn{2}{c}{UndeepVO \cite{c9}} & \multicolumn{2}{c}{Undepthflow\cite{c11}} \\
            
			& \multicolumn{2}{c}{(416$\times$128)}    & \multicolumn{2}{c}{(1242$\times$376)}                   & \multicolumn{2}{c||}{(416$\times$128)}                  & \multicolumn{2}{c||}{(1242$\times$376)}        & \multicolumn{2}{c}{(416$\times$128)} & \multicolumn{2}{c}{(832$\times$256)} \\
            
			& $t_{\text{rel}} {(\%)}$ & $r_{\text{rel}} (^{\circ})$          & $t_{\text{rel}} (\%)$ & $r_{\text{rel}} (^{\circ})$           & $t_{\text{rel}} (\%)$ & $r_{\text{rel}} (^{\circ})$ 		& $t_{\text{rel}} (\%)$ & $r_{\text{rel}} (^{\circ})$ & $t_{\text{rel}} (\%)$ & $r_{\text{rel}} (^{\circ})$ & $t_{\text{rel}} (\%)$ & $r_{\text{rel}} (^{\circ})$\\ \hline
            
			\multicolumn{1}{c|}{03}   & 10.56             & \multicolumn{1}{c|}{6.30} & \textbf{6.72}             & \multicolumn{1}{c|}{6.46}  & 13.08             & \multicolumn{1}{c||}{\textbf{3.79}}  & 0.67    & \multicolumn{1}{c||}{0.18}  & 5.00   & \multicolumn{1}{c|}{6.17}  & -      & -     \\
            
			\multicolumn{1}{c|}{04}   & \textbf{2.40}             & \multicolumn{1}{c|}{\textbf{0.77}} & 6.33             & \multicolumn{1}{c|}{6.08}  & 10.68            & \multicolumn{1}{c||}{5.13}  & {0.65}    & \multicolumn{1}{c||}{0.18}  & 4.49   & \multicolumn{1}{c|}{2.13}  & -      & -     \\
            
			\multicolumn{1}{c|}{05}   & \textbf{3.25}             & \multicolumn{1}{c|}{\textbf{1.31}} & 3.35             & \multicolumn{1}{c|}{4.93}  & 16.76             & \multicolumn{1}{c||}{4.06}  & 3.28    & \multicolumn{1}{c||}{0.46}  & 3.40   & \multicolumn{1}{c|}{1.50}  & -      & -     \\   
            
            \multicolumn{1}{c|}{06}   & \textbf{3.99}             & \multicolumn{1}{c|}{\textbf{1.46}} & 7.24             & \multicolumn{1}{c|}{7.29}  & 23.53             & \multicolumn{1}{c||}{4.80}  & 6.14    & \multicolumn{1}{c||}{0.17}  & 6.20   & \multicolumn{1}{c|}{1.98}  & -      & -     \\
            
            \multicolumn{1}{c|}{07}   & 4.67             & \multicolumn{1}{c|}{\textbf{1.83}} & \textbf{3.52}             & \multicolumn{1}{c|}{5.02}  & 17.52            & \multicolumn{1}{c||}{5.38}  & 1.23    & \multicolumn{1}{c||}{0.22}  & 3.15   & \multicolumn{1}{c|}{2.48}  & -      & -     \\ \hline

            \multicolumn{1}{c|}{$09^*$}   & \textbf{4.95}             & \multicolumn{1}{c|}{\textbf{2.37}} & {-}             & \multicolumn{1}{c|}{-}  & 18.77            & \multicolumn{1}{c||}{3.21}  & 15.30    & \multicolumn{1}{c||}{0.26}  & 7.01   & \multicolumn{1}{c|}{3.61}  &       13.98 & 5.36     \\  
            
			\multicolumn{1}{l|}{$10^*$}   & \textbf{5.89}             & \multicolumn{1}{c|}{\textbf{3.56}} & 9.77            & \multicolumn{1}{c|}{10.2}  & 14.36             & \multicolumn{1}{c||}{3.98}  & 3.68    & \multicolumn{1}{c||}{0.48}  & 10.63   & \multicolumn{1}{c|}{4.65}  & 19.67     & 9.13     \\            

		\end{tabular*}
		\begin{itemize}
			%\scriptsize
			\item $t_{\text{rel}}$: average translational RMSE drift $(\%)$ on length of 100m-800m.
			\item $r_{\text{rel}}$: average rotational RMSE drift $(^{\circ}/100m)$ on length of 100m-800m.
			\item $\text{train sequence}$: 03,04,05,06,07;
			$\text{test sequence}$: $09^*$, $10^*$
		\end{itemize}
	\end{table*}

\section{EXPERIMENTS}
We implemented the proposed SGANVO architecture with the publicly available Tensorflow framework and trained with NVIDIA GTX 1080TI GPUs. The Adam optimizer was employed to speed up the network convergence for up to 30 epochs with parameters $\beta_1 = 0.9$ and $\beta_2 = 0.999$. The
learning rate started from 0.0001 and decreased by half for every 1/5 of total iterations. For the ease of training and data preparation, we used a temporal window size of $N = 3$ and two higher layers $L=2$ and one bottom layer, but it is possible to use longer sequences and more higher layers for training and testing. The size of input image to the network was $416\times128$ with the consideration of comparison with other systems. To fine-tune the network, we used the original image size in computing the losses. For data prepossessing, different kinds of data augmentation methods were used to enhance the performance and mitigate possible over-fitting, such as image color augmentation\cite{c5}, rotational data augmentation\cite{c24} and left-right pose estimation augmentation\cite{c4}. We increased the weight parameter of rotational data to achieve better performance because the magnitude of rotation is very small compared to that of translation. We set $\alpha=10^{-4}$, $\beta=1.0$ and $\gamma=0.1$ in the final G loss \eqref{eq:7}. To test our SGANVO's depth estimation on the monocular dataset, we built a stereo view reconstruction component like \cite{c26} to generate the stereo frames from the monocular dataset ( see the testing inputs in Fig.1). Our SGANVO can test directly on the monocular dataset.

\begin{table}%[bp]% [thpb] % [h]
		\centering
		\renewcommand{\arraystretch}{1.45}
		\caption{ Absolute Trajectory Error (ATE) on KITTI odometry
dataset. The results of other baselines are taken from \cite{c11}.}
		\label{RobotCar_table}
		%\footnotesize
		% \scriptsize
		\small
		\setlength{\tabcolsep}{3.5pt}
		\begin{tabular*}{\columnwidth}{c|c|cc}
		\hline
		Method&frames & Sequence 09 & Sequence 10\\
		\hline
		ORB-SLAM(Full)&All&$0.014\pm0.008$&$0.012\pm0.011$\\
		SfMLearner \cite{c5} &$5$&$0.016\pm0.009$&$0.013\pm0.009$\\
		GeoNet \cite{c6}&$5$&$0.012\pm0.007$&$0.012\pm0.009$\\
		Undepthflow\cite{c11}&2&$0.023\pm0.010$&$0.022\pm0.016$\\
		%GANVO\cite{almalioglu2018ganvo}&2&$ 0.009\pm0.005$&$0.010\pm0.013$\\
		\hline
		SGANVO&$3$&$0.015\pm0.006$&$0.014\pm0.009$\\
		\hline
			
		\end{tabular*}
	\end{table}

\subsection{Depth Estimation Evaluation}
We evaluate the performance of our SGANVO depth estimation on the KITTI dataset with a benchmark split  from \cite{c5}. Fig. 3 shows some raw RGB images from sequence 10 and their corresponding depth estimates from
Monodepth \cite{c4}, and our system. As shown in Fig. 3, the different depths of cars and trees are explicitly generated, even the depth of trunks and street lights are generated clearly. Compared with the others, our SGANVO can generate more details and clearer textures.

The quantitative depth estimation results are listed in Table II, where $M$ stands for using monocular image sequence training and $S$ stands for using stereo image sequence training. Compared with the existing unsupervised learning-based methods (see the smallest values in Abs Rel, Sq Rel, RMSE, and RMSE log columns in the table with the highest $\delta$), our SGANVO performs better in terms of the metrics used. And it even outperforms some methods with larger image input.

\begin{table*}%[bp]% [thpb] % [h]
		\centering
		\renewcommand{\arraystretch}{3}
		\caption{Performance for different number of layers and window frame. NL = number of layers, NF = number of frames. Sq.09 and Sq.10 Absolute Trajectory Error(ATE) are the ego-motion testing results. From Abs Rel to $\delta$\textless $1.25^3$ are the depth testing results.}
		\label{spacial_temporal_comparison}
		\small
				\centering
		\renewcommand{\arraystretch}{1.45}
		\begin{tabular*}{1.0\textwidth}{c|c|cc|ccccccccc}
	%	\begin{tabular*}{\columnwidth}{ccccccccccc}
		\hline
		NL & NF & Sq.09 (ATE) & Sq.10 (ATE) & Abs Rel  & Sq Rel & RMSE  & RMSE $\log$&$\delta$ \textless 1.25 &$\delta$ \textless $1.25^2$&$\delta$ \textless $1.25^3$ \\ 
		\hline
        2  & 2  & $0.0178\pm0.0083$      & $0.0180\pm0.0107$     & 0.0991   & 0.873  & 4.483 & 0.184    & \textbf{0.902}            & \textbf{0.959}            & \textbf{0.979}             \\
        \hline
        2  & 3  & $0.0152\pm0.0067$      & $0.0149\pm0.0094$     & 0.0973   & 0.733  & 4.623 & 0.167    & \textbf{0.903}            & \textbf{0.969}            & \textbf{0.988}            \\
        \hline
        3  & 3  & $0.0153\pm0.0061$      & $0.0147\pm0.0085$     & 0.0651   & 0.673  & 4.003 & 0.136    & \textbf{0.944}            & \textbf{0.979}            & \textbf{0.991}             \\
        \hline
		\end{tabular*}
	\end{table*}

\subsection{Ego-motion Estimation Evaluation}
We used the KITTI dataset\cite{c1} to test the performance of the SGANVO ego-motion estimation. The KITTI dataset only provides the ground-truth of 6-DoF poses for Sequence 00-10. We used Sequence 00-08 for training and Sequence 09-10 for testing. 

The metrics are the average translational root-mean-square error (RMSE) drift and average rotational RMSE drift $(◦/100m)$ on length of $100m-800m$. The results are shown in Table III. We compare the results with ESP-VO \cite{c24}, SfMLearner\cite{c5}, ORB-SLAM (without loop closure), UndeepVO \cite{c9}, and Undepthflow\cite{c11}. It can be seen that our SGANVO shows a better performance in the testing sequences (09, 10) with these state-of-the-art methods in terms of the ATE metric.

The other metrics used are the absolute trajectory error (ATE) averaged over all overlapping 5-frame snippets. We concatenated all of left and right estimations together for the entire sequences without any post-processing. The estimated trajectory of sequences 9 and 10 from our SGANVO and its ground truth are shown in Fig. 4. Although the estimated result includes some drift, our SGANVO can estimate all the features of the trajectory and performs well in terms of odometry estimation without loop closure detection. The quantitative results are shown in Table IV where we compare our results with ORB-SLAM(Full), SfMLearner\cite{c5}, GeoNet \cite{c6}, and
Undepthflow\cite{c11}. Our SGANVO produced a comparable result.

\subsection{Ablation Analysis for Spatial Layers and Temporal Frames}
To demonstrate the importance of the high level
spatial features and temporal recurrent window, we conducted three contrast experiments for an ablation analysis. In  Table V, we set the same network and training parameters, but configured the system with different number of high level layers and size of temporal window to explore the influence of high level spatial features and temporal recurrent window on the learning results. The main evaluation is focused on the ego-motion and depth test for sequence 09 and sequence 10. Considering the GPU's memory and training time, we just tested the system configuration with 2 layers or 3 layers combining with 2 frames or 3 frames. 

%In order to better explain the problems, we keep a record of the more accurate data digit statistics. 
It can be seen from the data in the first two rows that the ego-motion performs much better when the window size increases from 2 to 3 when the number of high level layers are kept unchanged. In addition, the depth estimation also has a little bit improvement. Then we fixed the same temporal window size as 3 frames to explore the influence of different number of high level layers. It can be seen from the data in the last two rows of Table V that the number of high level layers can make a significant improvement on the depth estimation, although its influence on the ego-motion estimation is limited. 

From these results, we can conclude that the size of temporal recurrent window can make a significant improvement on the ego-motion estimation while the number of high level layers can make a significant improvement on the depth estimation. 

\section{CONCLUSIONS}
In this paper, we propose a novel stacked GAN network for depth estimation and ego-motion estimation from videos. Because the higher layers of our network can learn spatial features with different abstraction levels and the recurrent representation of our network can learn the temporal dynamics between consecutive frames, our SGANVO can generate more accurate depth estimation result and comparable ego-motion estimation result comparing with other existing unsupervised learning networks. It is the learning mechanism between the Generator and Discriminator that is able to improve the quality of the generated depth and ego-motion estimation. 

In the future, we will extend our system to a visual SLAM system to reduce the drift by adding a loop closure detection. And also more data sets will be used for training and testing to further improve the performance.


%\addcontentsline{toc}{section}{References}
%\bibliographystyle{IEEEtran}
%\bibliography{IEEEabrv,references}

\begin{thebibliography}{99}

\bibitem{c1} Geiger, Andreas, Philip Lenz, and Raquel Urtasun. "Are we ready for autonomous driving? the kitti vision benchmark suite." 2012 IEEE Conference on Computer Vision and Pattern Recognition. IEEE, 2012.
\bibitem{c2} Jaderberg, Max, Karen Simonyan, and Andrew Zisserman. "Spatial transformer networks." Advances in neural information processing systems. 2015.
\bibitem{c3} Garg, Ravi, et al. "Unsupervised cnn for single view depth estimation: Geometry to the rescue." European Conference on Computer Vision. Springer, Cham, 2016.
\bibitem{c4} Godard, Clément, Oisin Mac Aodha, and Gabriel J. Brostow. "Unsupervised monocular depth estimation with left-right consistency." Proceedings of the IEEE Conference on Computer Vision and Pattern Recognition. 2017.
\bibitem{c5} Zhou, Tinghui, et al. "Unsupervised learning of depth and ego-motion from video." Proceedings of the IEEE Conference on Computer Vision and Pattern Recognition. 2017.
\bibitem{c6} Yin, Zhichao, and Jianping Shi. "Geonet: Unsupervised learning of dense depth, optical flow and camera pose." Proceedings of the IEEE Conference on Computer Vision and Pattern Recognition. 2018.
\bibitem{c7} Godard, Clément, et al. "Digging into self-supervised monocular depth estimation." arXiv preprint arXiv:1806.01260 (2018).
\bibitem{c8} Pillai, Sudeep, Rares Ambrus, and Adrien Gaidon. "SuperDepth: Self-Supervised, Super-Resolved Monocular Depth Estimation." arXiv preprint arXiv:1810.01849 (2018).
\bibitem{c9} Li, Ruihao, et al. "Undeepvo: Monocular visual odometry through unsupervised deep learning." 2018 IEEE International Conference on Robotics and Automation (ICRA). IEEE, 2018.
\bibitem{c10} Ranjan, Anurag, et al. "Competitive Collaboration: Joint Unsupervised Learning of Depth, Camera Motion, Optical Flow and Motion Segmentation." Proceedings of the IEEE Conference on Computer Vision and Pattern Recognition. 2019.
\bibitem{c11} Wang, Yang, et al. "Joint Unsupervised Learning of Optical Flow and Depth by Watching Stereo Videos." arXiv preprint arXiv:1810.03654 (2018).
\bibitem{c12} Sun, Deqing, et al. "PWC-Net: CNNs for optical flow using pyramid, warping, and cost volume." Proceedings of the IEEE Conference on Computer Vision and Pattern Recognition. 2018.
\bibitem{c13} Almalioglu, Yasin, et al. "GANVO: Unsupervised Deep Monocular Visual Odometry and Depth Estimation with Generative Adversarial Networks." arXiv preprint arXiv:1809.05786 (2018).
\bibitem{c14} Goodfellow, Ian, et al. "Generative adversarial nets." Advances in neural information processing systems. 2014.
\bibitem{c15} CS Kumar, Arun, Suchendra M. Bhandarkar, and Mukta Prasad. "Monocular depth prediction using generative adversarial networks." Proceedings of the IEEE Conference on Computer Vision and Pattern Recognition Workshops. 2018.
\bibitem{c16} Aleotti, Filippo, et al. "Generative Adversarial Networks for unsupervised monocular depth prediction." Proceedings of the European Conference on Computer Vision (ECCV). 2018.
\bibitem{c17} Pilzer, Andrea, et al. "Unsupervised adversarial depth estimation using cycled generative networks." 2018 International Conference on 3D Vision (3DV). IEEE, 2018.
\bibitem{c18} Gwn Lore, Kin, et al. "Generative adversarial networks for depth map estimation from RGB video." Proceedings of the IEEE Conference on Computer Vision and Pattern Recognition Workshops. 2018. 
\bibitem{c19} Shi, Wenzhe, et al. "Real-time single image and video super-resolution using an efficient sub-pixel convolutional neural network." Proceedings of the IEEE conference on computer vision and pattern recognition. 2016.
\bibitem{c20} Klambauer, Günter, et al. "Self-normalizing neural networks." Advances in neural information processing systems. 2017.
\bibitem{c21} Eigen, David, Christian Puhrsch, and Rob Fergus. "Depth map prediction from a single image using a multi-scale deep network." Advances in neural information processing systems. 2014.
\bibitem{c22} Mahjourian, Reza, Martin Wicke, and Anelia Angelova. "Unsupervised learning of depth and ego-motion from monocular video using 3d geometric constraints." Proceedings of the IEEE Conference on Computer Vision and Pattern Recognition. 2018.
\bibitem{c23} Chavdarova, Tatjana, and François Fleuret. "SGAN: An Alternative Training of Generative Adversarial Networks." Proceedings of the IEEE Conference on Computer Vision and Pattern Recognition. 2018.
\bibitem{c24} Wang, Sen, et al. "End-to-end, sequence-to-sequence probabilistic visual odometry through deep neural networks." The International Journal of Robotics Research 37.4-5 (2018): 513-542.
\bibitem{c25} Mur-Artal, Raul, and Juan D. Tardós. "Orb-slam2: An open-source slam system for monocular, stereo, and rgb-d cameras." IEEE Transactions on Robotics 33.5 (2017): 1255-1262.
\bibitem{c26} Luo Y, Ren J, Lin M, et al. Single view stereo matching[C]//Proceedings of the IEEE Conference on Computer Vision and Pattern Recognition. 2018: 155-163.
\bibitem{c27} Cordts, Marius, et al. "The cityscapes dataset for semantic urban scene understanding." Proceedings of the IEEE conference on computer vision and pattern recognition. 2016.


\end{thebibliography}


\end{document}